\documentclass[runningheads]{llncs}
\usepackage{graphicx}
\usepackage[utf8x]{inputenc} 
\usepackage{amsmath}
\usepackage{bbm}
\usepackage{amsfonts}
\usepackage{multirow}
\usepackage{xcolor}
\usepackage[colorlinks=true,citecolor=blue,urlcolor=blue]{hyperref}
\usepackage{float} 
\usepackage{subfigure} 

\usepackage{comment}
\usepackage{color}
\usepackage[misc]{ifsym}
\usepackage{bbding}
\begin{document}
\title{FSSL-IRM: Federated Semi-supervised Learning in Medical Image Classification
through \\ Inter-client Relation Matching}
\title{Federated Semi-supervised Medical Image Classification
via Inter-client Relation Matching}
\titlerunning{Inter-client Relation Matching for Federated Semi-supervised Learning}
%
\author{Quande Liu\inst{1}  
        \and Hongzheng Yang \inst{2} 
        \and Qi Dou \inst{1}
        \and Pheng-Ann Heng\inst{1,3}}  
\institute{Department of Computer Science and Engineering, The Chinese University of Hong Kong, Hong Kong SAR, China\\\email{\{qdliu, qdou\}@cse.cuhk.edu.hk} \and Department of Computer Science and Engineering, Beihang University, China \and Shenzhen Key Laboratory of Virtual Reality and Human Interaction Technology, Shenzhen Institutes of Advanced Technology, Chinese Academy of Sciences, China}%
%
\maketitle              
\vspace{-3mm}
\begin{abstract}
Federated learning (FL) has emerged with increasing popularity to collaborate distributed medical institutions for training deep networks. 
However, despite existing FL algorithms only allow the supervised training setting, most hospitals in realistic usually cannot afford the intricate data labeling due to absence of budget or expertise. 
This paper studies a practical yet challenging FL problem, named \textit{Federated Semi-supervised Learning} (FSSL), which aims to learn a federated model by jointly utilizing the data from both labeled and unlabeled clients (i.e., hospitals). We present a novel approach for this problem, which improves over traditional consistency regularization mechanism with a new inter-client relation matching scheme. The proposed learning scheme explicitly connects the learning across labeled and unlabeled clients by aligning their extracted disease relationships,  thereby mitigating the deficiency of task knowledge at unlabeled clients and promoting discriminative information from unlabeled samples.
We validate our method on two large-scale medical image classification datasets. The effectiveness of our method has been demonstrated with the clear improvements over state-of-the-arts as well as the thorough ablation analysis on both tasks\footnote{Code will be made available at \url{https://github.com/liuquande/FedIRM}}. 
\end{abstract}
\vspace{-8mm}
\section{Introduction}

Data collaboration across medical institutions is increasingly desired to mitigate the scarcity and distribution bias of medical images, thereby improving the model performance on important tasks such as disease diagnosis~\cite{dhruva2020aggregating,silva2019federated}. 
Recently, federated learning (FL) has emerged as a privacy-preserving solution for this, which allows to learn from distributed data sources  by aggregating the locally learned model parameters without exchanging the sensitive health data~\cite{dou2021federated,kaissis2020secure,li2019privacy,liu2021feddg,rieke2020future}. However, despite progress achieved, existing FL algorithms typically only allow the supervised training setting~\cite{chang2020synthetic,li2020federated,li2020multi,roth2020federated,sheller2018multi}, which has limited the local clients (i.e., hospitals) without data annotations to join the FL process.  
Yet, in realistic scenarios, most hospitals usually cannot afford the intricate data labeling due to lack of budget or expertise~\cite{razzak2018deep}. 
How to utilize these widely-existing unlabeled datasets to further improve the FL models is still an open question to be solved.

To this end, we study a practical FL problem which involves only several labeled clients while most of the participanting clients are unlabeled, namely \textit{federated semi-supervised learning (FSSL)}. 
This is also noted by Yang et al.~\cite{yang2021federated} very recently in COVID-19 lesion segmentation, which  halts in an extremely simple case containing only one labeled and one unlabeled client. 
In contrast to their work, we for the first time broaden this problem to a more practical yet complex scenario in which multiple distributed labeled and unlabeled clients are involved. 
To address this problem, a naive solution is to simply integrate the off-the-rack semi-supervised learning (SSL) methods onto the federated learning paradigm. 
However, previous SSL methods are typically designed for centralized training setting~\cite{bai2017semi,gyawali2020semi,liu2020semi,wang2020focalmix}, which \textit{rely heavily on the assumption that the labeled data is accessible to provide necessary assistance for the learning from  unlabeled data}~\cite{aviles2019graph,cheplygina2019not}. 
In consistency-based methods~\cite{cui2019semi,yu2019uncertainty}, for instance, the regularization of perturbation-invariant model predictions needs the synchronous labeled data supervision, in order to obtain the necessary task knowledge to produce reliable model predictions for unlabeled data where the consistency regularization is imposed on. 
Unfortunately, such close assistance from labeled data is lost in FSSL scenario, where the local dataset could be completely unlabeled. This will make the local model aloof from original task as the consistency-based
training goes on, hence fail to fully exploit the knowledge at unlabeled clients.
%

\vspace{-1mm}
Based on the above issues, compared with traditional SSL problem, the main challenge in FSSL lies in how to build the interaction between the learning at labeled and unlabeled clients, given the challenging constraint of data decentralization. In this work, our insight is to communicate their knowledge inherent in disease relationships to achieve this goal. The idea is motivated by an observation that the relationships exist naturally among different categories of disease and reflect the structural task knowledge in the context of medical image classification, as evidenced by disease similarity measure~\cite{mathur2012finding,oerton2019understanding}. More importantly, such disease relationships are independent of the observed hospitals, i.e., similar disease at one hospital should also be high-related at others. We may consider extracting such client-invariant disease relation information from labeled clients to supervise the learning at unlabeled clients, thereby mitigating the loss of task knowledge at unlabeled clients and effectively exploiting the unlabeled samples.

\vspace{-1mm}
In this paper, we present \textit{to our knowledge the first FSSL framework} for medical image classification, by exploring the client-independent disease relation information to facilitate the learning at unlabeled clients. Our method roots in the state-of-the-art consistency regularization mechanism, which enforces the prediction consistency under different input perturbations to exploit the unlabeled data. To address the loss of task knowledge at unlabeled clients which would lead to degenerated learning, we introduce a novel \textit{Inter-client Relation Matching} scheme, by explicitly regularizing the unlabeled clients to capture similar disease relationships as labeled clients for preserving the discriminative task knowledge. To this end, we propose to derive the disease relation  matrix  at labeled clients from pre-softmax features, and devise an uncertainty-based scheme to estimate reliable relation matrix at unlabeled clients by filtering out inaccurate pseudo labels. 
We validate our method on two large-scare medical image classification datasets, including intracranial hemorrhage diagnosis with 25000 CT slices and skin lesion diagnosis with 10015 dermoscopy images. Our method achieves large improvements by utilizing the unlabeled clients, and clearly outperforms the combination of federated learning with state-of-the-art SSL methods. 
\section{Method}
\begin{figure}[t]
	\centering
	\includegraphics[width=0.99\textwidth]{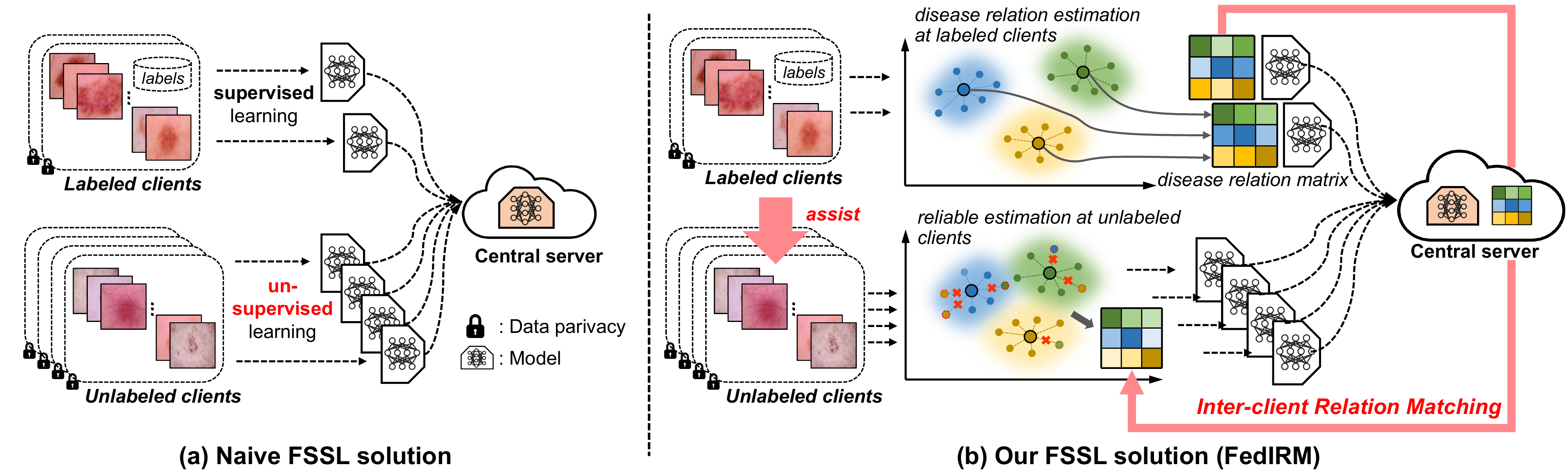}
	\vspace{-3mm}
	\caption{(a) Naive FSSL solution simply performs unsupervised learning (e.g. consistency regularization) at unlabeled clients, hence the local model is prone to forget the original task knowledge as the training goes on. (b) Our FedIRM explicitly utilizes the knowledge from labeled clients to assist the learning at unlabeled clients by aligning their extracted disease relationships, thereby mitigating the loss of task knowledge at unlabeled clients and promoting discriminative information from unlabeled data.}
	\vspace{-4mm}
	\label{fig:method}
\end{figure}

In FSSL scenario, we denote $\mathcal{D}_{L} = \{\mathcal{D}^{1}, \mathcal{D}^{2}, \dots, \mathcal{D}^{m}\}$ be the collection of $m$ labeled clients, where each labeled client $l$ contains $N^l$ data and one-hot label pairs $\mathcal{D}^{l} = \{(x_i^l, y_i^l)\}_{i=1}^{N^l}$; and let $\mathcal{D}_U = \{\mathcal{D}^{m+1}, \mathcal{D}^{m+2}, \dots, \mathcal{D}^{m+n}\}$ be the $n$ unlabeled clients, with each unlabeled client $u$ containing $N^u$  data samples $\mathcal{D}^{u} = \{(x_i^u)\}_{i=1}^{N^u}$. The goal of FSSL is to learn a global federated model $f_{\theta}$ jointly utilizing the data from both labeled and unlabeled clients. Fig.~\ref{fig:method} gives an overview of our proposed FSSL solution, i.e. FedIRM, in comparison with the naive FSSL solution.
\vspace{-1mm}
\subsection{Backbone Federated Semi-supervised Learning Framework}

Our method follows standard FL paradigm which involves the communication between a central server and local clients. Specifically, in each federated round, every client $k$ will receive the same global model parameters $\theta$ from the central server, and update the model with local learning objective $\mathcal{L}^k$ for $e$ epochs on its private data $\mathcal{D}^{k}$ . The central server then collects the local model parameters $\theta^k$ from all clients and aggregate them to update the global model. Such process repeats until the global model converges stably. In this work, we adopt the well-established federated averaging algorithm~\cite{mcmahan2017communication} (FedAvg) to update the global model, by aggregating the local model parameters with weights in proportional to the size of each local dataset, i.e., $\theta = \sum_{k=1}^{K} \frac{N^k}{N} \theta^k$, where $N=\sum_{k=1}^{K}N^k$.

In our FSSL solution, the local learning objective at labeled clients adopts the cross entropy loss for capturing the discriminative task knowledge. At unlabeled clients, we preserve the state-of-the-art consistency regularization mechanism, which exploits the unlabeled data in an unsupervised manner by enforcing the consistency of model predictions under input perturbations. Formally, this learning objective at each unlabeled client $u$ could be expressed as:
\begin{equation}
\vspace{-1mm}
    \mathcal{L}_c (\mathcal{D}^u, \theta^u) = \sum_{i=1}^{N^u} \mathbb{E}_{\xi, \xi'} ||f_{\theta^u}(x^u_i, \xi), f_{\theta^u}(x^u_i, \xi')||_2^2
    \label{eq:consistency}
\end{equation}
where $\xi$ and $\xi'$ denote different input perturbations (e.g., adding Gaussian noise). 








\subsection{Disease Relation Estimation at Labeled and Unlabeled Clients}
Without the assistance from labeled data supervision, the local learning at unlabeled clients solely with consistency regularization is prone to forget the original task knowledge, therefore failing to fully exploit the information from unlabeled samples. To tackle this problem, we introduce a novel \textit{inter-client relation matching} (IRM) scheme, which explicitly extracts the knowledge from labeled clients to assist the learning at unlabeled clients, by exploiting the rich information inherent in disease relationships.  Specifically, the relationships exist naturally across different categories of disease and reflect structural task knowledge in medical image classification, \textit{independent of the changes of observed hospitals}. In light of this, we aim to enforce the alignment of such disease relations across labeled and unlabeled clients, thereby promoting the learning of discriminative information at unlabeled clients for preserving such structural task knowledge.

\vspace{-3mm}
\subsubsection{Disease Relation Estimation at Labeled Clients.}
Inspired by knowledge distillation from deep networks, we estimate the disease relationships from the class ambiguity captured by deep models, i.e., per-class soft labels, and enforce them to be consistent between labeled and unlabeled clients. Formally, we first consider the relation estimation at labeled clients. For each labeled client $\mathcal{D}^l$, we summarize the model's knowledge on each class $c$ by computing per-category mean feature vectors $\mathbf{v}^l_{c}\in\mathbb{R}^{C}$ (with $C$ denoting total class number):
 \vspace{-1mm}
\begin{equation}
    \vspace{-1mm}
    \mathbf{v}^l_{c} = \frac{1}{N^l_{c}}  \sum_{i=1}^{N^l} \mathbbm{1}_{[y^l_i=c]} \hat{f}_{\theta^l} (x^l_i) 
    \label{eq:mean_feature_vection}
\end{equation}
where $N^l_{c}$ is the number of samples with class $c$ at labeled client $\mathcal{D}^l$; $\mathbbm{1}_{[\cdot]}$ denotes the indicator function, $\hat{f}$ denotes the model without last softmax layer. The obtained $\mathbf{v}^l_{c}$ is then scaled to a soft label distribution, with a softened softmax function under temperature $\tau>1$~\cite{dou2020unpaired}:
 \vspace{-1mm}
\begin{equation}
    \mathbf{s}^l_{c} = \textnormal{softmax}(\mathbf{v}^l_{c}/\tau)
    \label{eq:softmax}
\end{equation}
This distilled knowledge of soft label $\mathbf{s}^l_{c}$ conveys how the network predictions of samples on certain class generally distribute across all classes, reflecting the relationships across different classes captured by the deep model. Consequently, the collection of soft labels from all classes could form a soft confusion matrix $\mathcal{M}^l = [\mathbf{s}^l_{1}, \dots, \mathbf{s}^l_{C}]$, which encodes the inter-class relationships among different categories of disease hence serve as the disease relation matrix. 
\vspace{-2mm}
\subsubsection{Reliable Disease Relation Estimation at Unlabeled Clients.} Since the data annotations are unavailable at unlabeled clients, we utilize the pseudo labels generated from model predictions to estimate the disease relation matrix. However, without sufficient task knowledge provided at unlabeled clients, the model predictions on unlabeled data could be noisy and inaccurate. We therefore employ an uncertainty-based scheme to filter out the unreliable model predictions, and only preserve the trustworthy ones to measure the reliable relation matrix.

Specifically, we take the local training at unlabeled client $\mathcal{D}^u$ for instance. Given an input mini-batch $\mathbf{x}^u$ of $B$ image, we denote $\mathbf{p^u}$ as the corresponding predicted probability and $\mathbf{y^u}$ as the pseudo labels, i.e., $\mathbf{y^u} = \textnormal{argmax} (\mathbf{p^u})$. Following the literature on uncertainty estimation, we approximate the uncertainty of model predictions with dropout of Bayesian networks~\cite{gal2016dropout}. Concretely, we perform $T$-time forward propagation for the input mini-batch $\mathbf{x}^u$ under random dropout, obtaining a set of predicted probability vectors $\{\mathbf{q}^u_t\}_{t=1}^T$. The uncertainty $\mathbf{w}^u$ is then estimated as the predictive entropy, which is computed from the averaged probability from the $T$-time forward passes as: 
\vspace{-2mm}
\begin{equation}
\vspace{-2mm}
    \mathbf{w}^u=-\sum_{c=1}^{C} \overline{\mathbf{q}}_{(c)}^u ~\textnormal{log}(\overline{\mathbf{q}}_{(c)}^u), ~\textnormal{with}~ \overline{\mathbf{q}}_{(c)}^u = \frac{1}{T} \sum_{t=1}^{T} \mathbf{q}^u_{t(c)}
\end{equation}
where $\mathbf{q}_{t(c)}^u$ is the value of the $c$-th class of $\mathbf{q}_{t}^u$.
Since the predictive entropy has a fixed range, we can filter out the relatively unreliable predictions and only select the certain ones to compute the disease relation matrix. Hence, the per-category mean feature vectors $\mathbf{v}^u_c$ (c.f. Eq.~\ref{eq:mean_feature_vection}) at unlabeled clients are computed as:
\vspace{-1mm}
\begin{equation}
    \mathbf{v}^u_{c} = \frac{\sum_{i=1}^{B} \mathbbm{1}_{[(\mathbf{y}_i=c) \cdot (\mathbf{w}^u_i < h)]}  \cdot \mathbf{p}^u_i}{\sum_{i=1}^{B} \mathbbm{1}_{[(\mathbf{y}_i=c) \cdot (\mathbf{w}^u_i < h)]}}
\end{equation}
where $h$ is the threshold to select the certain predictions from $\mathbf{w}^u$. Then, following the same operation as Eq.~\ref{eq:softmax}, the disease relation matrix at unlabeled client $u$ is estimated as $\mathcal{M}^u = [\mathbf{s}^u_{1}, \mathbf{s}^u_{2}, \dots, \mathbf{s}^u_{C}]$. 
\vspace{-3mm}
\subsection{Objective of Inter-client Relation Matching}
\vspace{-1mm}
With the above basis, we aim to enforce the unlabeled clients to produce similar  disease relationships as labeled clients to preserve such discriminative task knowledge. Specifically, at the end of each federated round, the central server collects the relation matrix $\mathcal{M}^l$ from each labeled client, and average them to compute a matrix representing the general disease relation information captured from all labeled data, i.e., $\mathcal{M} = \frac{1}{m} \sum_{l=1}^{m} \mathcal{M}^l$. This obtained $\mathcal{M}$ is then delivered to unlabeled clients to supervise their next round of local training. To establish the supervision online, the relation matrix at unlabeled clients $\mathcal{M}^u$ is estimated from each mini-batch during training. Finally, the inter-client relation matching loss is designed by minimizing the KL divergence between $\mathcal{M}$ and $\mathcal{M}^u$ as:
\vspace{-2mm}
\begin{equation}
\begin{aligned}
    \mathcal{L}_{\textnormal{IRM}} &= \frac{1}{C} \sum_{c=1}^{C}( \mathcal{L}_{\textnormal{KL}} (\mathcal{M}_c || \mathcal{M}_c^u) + \mathcal{L}_{\textnormal{KL}} (\mathcal{M}_c^u || \mathcal{M}_c)),\\ &~\textnormal{with} ~ \mathcal{L}_{\textnormal{KL}} (\mathcal{M}_c || \mathcal{M}_c^u)) = \sum_{j} \mathcal{M}_{{c(j)}} \textnormal{log} \frac{\mathcal{M}_{c(j)}}{\mathcal{M}_{c(j)}^u}
    \label{eq:FedIRM}
    \vspace{-4mm}
\end{aligned}
\end{equation}
where $\mathcal{M}_c \in \mathbb{R}^{C}$ denote the relation vector of class $c$, i.e., $\mathcal{M}_c = \mathbf{s}_{c}$; and $\mathcal{M}_{c(j)}$ denote its $j$-th entry.
Overall, the local learning objectives at labeled ($\mathcal{L}^{l}$) and unlabeled ($\mathcal{L}^{u}$) clients are respectively expressed as:
\begin{equation}
\begin{aligned}
\vspace{-1mm}
    \mathcal{L}^{l} = \mathcal{L}_{ce} (\mathcal{D}^l, \theta^l) ~~\textnormal{and} ~~
    \mathcal{L}^{u} = \lambda(\omega) (\mathcal{L}_{c} + \mathcal{L}_\textnormal{IRM})
\end{aligned}
\end{equation}
where $\mathcal{L}_{ce}$ is the cross entropy loss; $\mathcal{L}_{c}$ is the traditional consistency regularization loss (c.f. Eq.~\ref{eq:consistency}); $\lambda(\omega)$ is a warming up function regarding federated round $\omega$, which helps to reduce the effect of the learning at unlabeled clients when the model is underfitting at earlier federated rounds.

\vspace{-3mm}
\section{Experiments}
\vspace{-1mm}
\subsection{Dataset and Experimental Setup} 
\vspace{-1mm}
We validate our method on two important medical image classification tasks, including: intracranial hemorrhage (ICH) diagnosis from brain CT and skin lesion classification  from dermoscopy images. 

\textbf{Task 1 - Intracranial hemorrhage diagnosis.} We perform ICH diagnosis with the RSNA ICH Detection dataset\cite{rsna2019ichchallenge}, which aims to classify CT slices into 5 subtypes of ICH disease. Since most images in this dataset are healthy without any of the subtypes, we randomly sample 25000 slices from the dataset which contain one of the 5 subtypes of ICH for evaluation. These samples are then randomly divided into 70\%, 10\% and 20\% for training, validation and testing. Since multiple slices may come from the same patient in this dataset, we have ensured no overlapped patients exist across the three split for a valid evaluation. 

\textbf{Task 2 - Skin lesion diagnosis}. We employ ISIC 2018: Skin Lesion Analysis Towards Melanoma Detection\cite{tschand2018ham} dataset for skin lesion diagnosis, which contains 10015 dermoscopy images in the official training set labeled by 7 types of skin lesions. As the ground truth of official validation and testing set was not released, we randomly divide the entire training set to 70\% for training, 10\% for validation and 20\% for testing. We perform the same data pre-processing for the two tasks. Specifically, we first resized the original images from 512 × 512 to 224 × 224. To employ the pre-trained model, we then normalized the images with statistic collected from ImageNet dataset before feeding them into the network. 


\textbf{Experiment Setup.} To simulate the FL setting, we randomly partition the training set into 10 different subsets serving as 10 local clients. Following the practice in SSL~\cite{tarvainin2017mean}, we evaluate the model performance under 20\% labeled data setting, i.e., two clients are labeled and the remaining eight are unlabeled in our case. 
Five metrics are used to extensively evaluate the classification performance, including AUC, Sensitivity, Specificity, Accuracy and F1 score. We report the results in form of average and standard deviation over three independent runs. 
\vspace{-2mm}
\subsubsection{Implementation Details.}
We employ DenseNet121~\cite{huang2017densely} as the backbone for medical image classification. Two types of perturbations are utilized to drive the consistency regularization, including random transformation on input data (rotation, translation and flip) and dropout layer in the network. 
The temperature parameter $\tau$ is empirically set as 2.0. The forward pass time $T$ used to compute uncertainty is set as 8, and the threshold $h$ to select reliable predictions is set as ln2. We follow~\cite{tarvainin2017mean} to apply a Gaussian warming up function $\lambda(\omega) = 1*e^{(-5(1-\omega/\Omega))}$, where $\Omega$ is set as 30. The local training adopts Adam optimizer with momentum of 0.9 and 0.99, and the batch size is  48 for both labeled and unlabeled clients. We totally train 100 federated rounds when the global model has converged stably, with the local training epoch $e$ set as 1. 
\begin{table}[t]
    \renewcommand\arraystretch{1.1}
	\centering
    \caption{Quantitative comparisons with state-of-the-arts on two different tasks.}
    \vspace{-3mm}
    \label{tab:tab1}
    \scalebox{0.83}{
    \begin{tabular}{c|cc|c|c|c|c|c}
        \hline
        \hline
        \multirow{2}{*}{Method} & \multicolumn{2}{c|}{ \textbf{Client num}} &\multicolumn{5}{c}{\textbf{Metrics}} \\
        \cline{2-8}
                    &Label  &Unlabel    &AUC     &Sensitivity    &Specificity   &Accuracy    &F1    \\
        \hline
        \hline
        &\multicolumn{7}{c}{\textbf{Task 1: Intracranial hemorrhage diagnosis}}\\
        \hline
        {FedAvg~\cite{mcmahan2017communication}} &10   &0  &90.48$\pm$0.31  &64.33$\pm$1.13  &92.68$\pm$0.43 &89.94$\pm$0.92 &63.94$\pm$1.20   \\
        \hline 
    {FedAvg~\cite{mcmahan2017communication}}
                    &2   &0    &83.40$\pm$0.87 &57.88$\pm$1.68 &90.48$\pm$0.79 &87.45$\pm$1.08 &57.10$\pm$1.29    \\
      Fed-SelfTraining~\cite{zhang2020benchmarking}
                    &2   &8  &84.32$\pm$0.82 &57.94$\pm$1.66 &90.22$\pm$0.74 &87.90$\pm$1.81 &57.48$\pm$1.14\\
      {Fed-Consistency~\cite{yang2021federated}}
                    &2   &8     &84.83$\pm$0.79     &57.26$\pm$1.93 &90.87$\pm$0.62  &88.35$\pm$1.32  &57.61$\pm$1.08    \\
        \hline
      {\textbf{FedIRM (ours)}}
                    &2   &8     &\textbf{87.56$\pm$0.56} &\textbf{59.57$\pm$1.57} &\textbf{91.53$\pm$0.81} &\textbf{88.89$\pm$1.29} &\textbf{59.86$\pm$1.65}  \\ 
        \hline
        \hline
    &\multicolumn{7}{c}{\textbf{Task 2: Skin Lesion Diagnosis}}\\
    \hline
    {FedAvg~\cite{mcmahan2017communication} }
                    &10   &0    &94.82$\pm$0.32 &75.11$\pm$1.82 &94.87$\pm$0.35 &95.24$\pm$0.21 &70.16 $\pm$1.21\\
                    \hline
    {FedAvg~\cite{mcmahan2017communication}}
                    &2   &0   &90.65$\pm$1.23 &65.53$\pm$1.76 &91.76$\pm$0.48 &92.53$\pm$0.67 &52.59$\pm$1.42 \\
      Fed-SelfTraining~\cite{zhang2020benchmarking}
                    &2   &8 &90.82$\pm$0.56 &67.03$\pm$1.93 &\textbf{93.61$\pm$0.21} &92.47$\pm$0.34 &53.44$\pm$1.85 \\
      {Fed-Consistency~\cite{yang2021federated}}
                    &2   &8  &91.13$\pm$0.62 &68.55$\pm$1.29 &93.45$\pm$0.94 &92.67$\pm$0.39 &54.25$\pm$1.31     \\
        \hline
      {\textbf{FedIRM (ours)}}
                    &2   &8  &\textbf{92.46$\pm$0.45} &\textbf{69.05$\pm$1.71} &93.29$\pm$0.59 &\textbf{92.89$\pm$0.25} &\textbf{55.81$\pm$1.49}    \\ 
    \hline
    \hline
    \end{tabular}
    }
    \vspace{-5mm}
    \label{tab:comparison}
\end{table}
\subsection{Comparison with State-of-the-arts.}  
\vspace{-1mm}
We compare with recent FSSL methods, including \textbf{Fed-SelfTraining}\cite{zhang2020benchmarking}, which performs self training at unlabeled clients by iteratively updating the model parameters and the pseduo labels of unlabeled data with expectation maximization; and  \textbf{Fed-Consistency}~\cite{yang2021federated}, which employs the state-of-the-art SSL strategy, i.e., consistency regularization, to exploit the data at unlabeled clients \textit{(without inter-client relation matching compared with our method)}. We also compare with the \textbf{FedAvg}~\cite{mcmahan2017communication} model trained only with labeled clients or with all clients as labeled, which serve as the baseline and upperbound performance in FSSL. 

The results on the two tasks are listed in Table~\ref{tab:comparison}. As observed, both Fed-SelfTraining and Fed-Consistency performs better than the baseline FedAvg model, which reflects the benefit to integrate the knowledge from additional unlabeled clients to improve FL models. Notably, compared with these methods, our FedIRM achieves higher performance on nearly all metrics, with 2.73\% and 1.33\% AUC improvements  on the two tasks over Fed-Consistency which does not employ our inter-client relation matching scheme. These clear improvements benefit from our FedIRM scheme which explicitly harnesses the discriminative relation information  learned from labeled clients to facilitate the learning at unlabeled clients. Without the supervision from $\mathcal{L}_\textnormal{IRM}$, the local training simply from consistency regularization is prone to forget the original task information, hence fail to fully exploit the discriminative information from unlabeled data.


   

\vspace{-3mm}
\begin{figure}[t]
	\centering
	\includegraphics[width=0.99\textwidth]{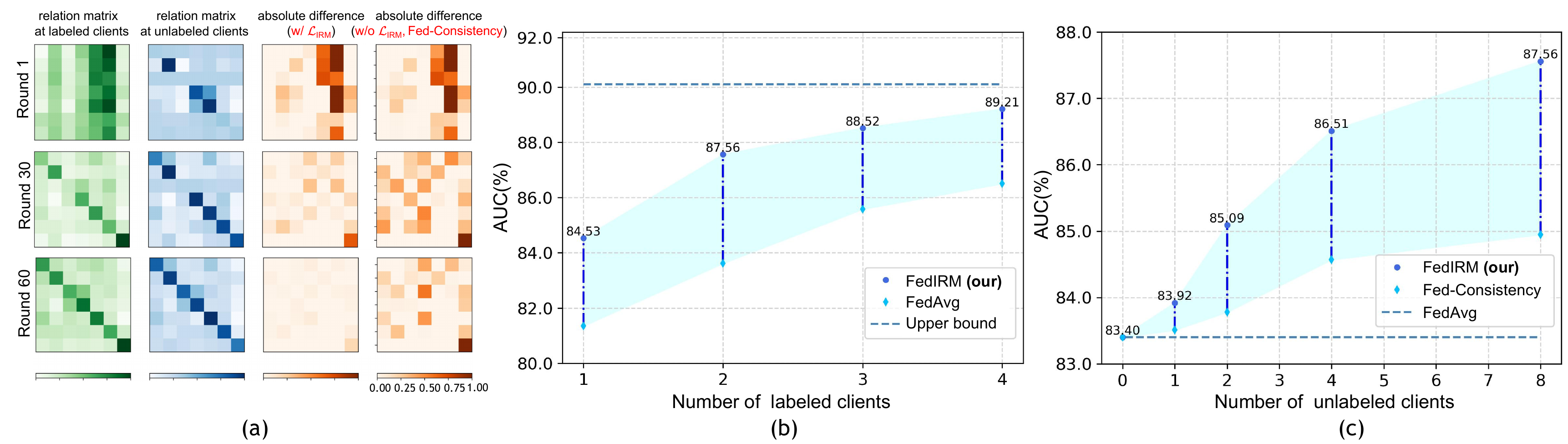}
	\vspace{-5mm}
	\caption{Ablation analysis. (a) Disease relation matrix at labeled and unlabeled clients under our method, as well as their absolute difference with or without $\mathcal{L}_\textnormal{IRM}$ (task 2); (b) Model performance under different labeled client number setting, using our approach and FedAvg (task 1); (c) Model performance as the number of unlabeled client increases (with labeled client number fixed), using our approach and Fed-Consistency (task 1).}
	\label{fig:ablation}
	\vspace{-5mm}
\end{figure}
\vspace{-1mm}
\subsection{Analytical Studies of Our Method}
\vspace{-1mm}
\subsubsection{Learning behavior under Inter-client Relation Matching.}
Fig.~\ref{fig:ablation}(a) displays the disease relation matrix of labeled (first col.) and unlabeled clients (second col.) under our method, as well as their absolute difference under our method with (third col.) and without $\mathcal{L}_\textnormal{IRM}$ (i.e., Fed-Consistency, forth col.), at different federated rounds. As observed, the relationships between disease at labeled clients become increasingly clear as the federated training goes on, indicating that the model gradually captures such structural knowledge. Notably, the unlabeled clients in our method can well preserve such disease relationships, with highly consistent matrix patterns as labeled clients and low responses in the difference matrix. In contrast, the method without $\mathcal{L}_\textnormal{IRM}$ (i.e., Fed-Consistency) fails to do so and the responses in difference matrix are relatively high. This observation affirms the benefit to transfer such discriminative knowledge to facilitate the learning at unlabeled clients and also explains our performance gains.

\vspace{-4mm}
\subsubsection{Effectiveness under different labeled client number.} We investigate the impact of different labeled client number in our method. As shown in the curve of Fig.~\ref{fig:ablation}(b), our method shows consistent improvements over the supervised-only FedAvg model under labeled client number from 1 to 4 (corresponds 10\% to 40\% labeled data setting in SSL). Importantly, using only 40\% labeled client, our method achieves 89.21\% AUC, which is very close to the upper-bound FedAvg model trained with 10 labeled clients (90.48\%). This endorses the capability of our method to leverage the data from unlabeled clients for improving FL models.
\vspace{-8mm}
\subsubsection{Effect of adding more unlabeled clients.}
We finally analyze the effect of unlabeled client number on the performance of our FSSL method and Fed-Consistency, by fixing the labeled client number as 2 and gradually increasing the unlabeled client number in [1, 2, 4, 8]. As shown in Fig.~\ref{fig:ablation}(c), an interesting finding is the FSSL performance progresses as the unlabeled client number increases, indicating the potential in realistic scenarios to aggregate more widely-existing unlabeled clients to improve the FL models. Notably, our method consistently outperforms the Fed-Consistency method under different unlabeled client number, highlighting the stable capacity of our proposed FSSL learning scheme.



\vspace{-3mm}
\section{Conclusion}
\vspace{-3mm}
We present a new FSSL framework, which to our knowledge is the first method incorporating unlabeled clients to improve FL models for medical image classification. To address the deficiency of consistency regularization in FSSL, our method includes a novel inter-client relation matching scheme to explicitly utilize the knowledge of labeled clients to assist the learning at unlabeled clients. Experiments on two large-scare datasets demonstrate the effectiveness. Our method is extendable to non-IID scenario in FSSL setting, as the employed disease relations are independent of the observed clients and unaffected by image distributions.
\section{Acknowledgement}
The work described in this paper was supported in parts by the following grants:
Key-Area Research and Development Program of Guangdong Province, China (2020B010165004),
Hong Kong Innovation and Technology Fund (Project No. GHP/110/19SZ),
Foundation of China with Project No. U1813204.

\bibliographystyle{splncs04}
\bibliography{refs}
\end{document}


\title{Supplementary File}
\author{Paper ID: 299}
\institute{}

\maketitle

\begin{table}
    \renewcommand\arraystretch{1.1}
	\centering
    \caption{We conduct ablation study for the effect of each component in the proposed method on ICH diagnosis task. Utilizing consistency regularization to exploit unlabeled clients improves 1.43\% for the AUC score. On top of this, enforcing the alignment of disease relation matrix to facilitate the learning at unlabeled clients leads to a larger increase of 2.08\% AUC. Finally, filtering out the inaccurate pseudo labels to promote reliable relation estimation at unlabeled clients completes our method, and shows consistent improvements on different evaluation metrics.}
    \vspace{-1mm}
    \label{tab:tab1}
    \begin{tabular}{c|ccccc}
        \hline
        \hline
        \multirow{2}{*}{Method}  &\multicolumn{5}{c}{Metrics} \\
        \cline{2-6}
                    &AUC     &Sensitivity    &Specificity   &Accuracy    &F1    \\
        \hline
    {FedAvg}
   
                      &83.40 &57.88 &90.48 &87.45 &57.10    \\
     \hline
        + Consistency Regularization &84.83     &57.26 &90.87  &88.35  &57.61 \\
        + Inter-client Relation Matching &86.91 &58.18 &{91.48} &88.62 &57.75 \\
        + Reliable Relation Estimation&\textbf{87.56} &\textbf{59.57} &\textbf{91.53} &\textbf{88.89} &\textbf{59.86}  \\
        \hline
    \hline
    \end{tabular}
    \label{tab:ablation}
    \vspace{-7mm}
\end{table}